\documentclass{article}

\usepackage{arxiv}

\usepackage[utf8]{inputenc} 
\usepackage[T1]{fontenc}    
\usepackage{hyperref}       
\usepackage{url}            
\usepackage{booktabs}       
\usepackage{amsfonts,amsmath}       
\usepackage{nicefrac}       
\usepackage{microtype}      
\usepackage{lipsum}		
\usepackage{graphicx}
\usepackage{natbib}
\usepackage{doi}
\usepackage{listings}
\lstdefinelanguage{PDDL}
{
  sensitive=false,    
  morecomment=[l]{;}, 
  alsoletter={:,-},   
  morekeywords={
    define,domain,problem,not,and,or,when,forall,exists,either,
    :domain,:requirements,:types,:objects,:constants,
    :predicates,:action,:parameters,:precondition,:effect,
    :fluents,:primary-effect,:side-effect,:init,:goal,
    :strips,:adl,:equality,:typing,:conditional-effects,
    :negative-preconditions,:disjunctive-preconditions,
    :existential-preconditions,:universal-preconditions,:quantified-preconditions,
    :functions,assign,increase,decrease,scale-up,scale-down,
    :metric,minimize,maximize,
    :durative-actions,:duration-inequalities,:continuous-effects,
    :durative-action,:duration,:condition
    ,:inputs,:outputs,:certified
  }
}

\title{Maneuver Decision-Making with Trajectory Streams Prediction for Autonomous Vehicles}

\author{ \href{https://orcid.org/0000-0002-9662-0858}{\includegraphics[scale=0.06]{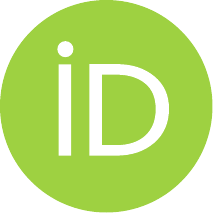}\hspace{1mm}Mais Jamal}\\
	Cognitive Modeling Center\\
	Moscow Institute of Physics and Technology \\
	141701 Dolgoprudny, Russia \\
	\texttt{mayssjamal@phystech.edu} \\
	\And
	\href{https://orcid.org/0000-0002-9747-3837}{\includegraphics[scale=0.06]{orcid.pdf}\hspace{1mm}Aleksandr Panov} \\
	Cognitive Modeling Center\\
	Moscow Institute of Physics and Technology \\
	141701 Dolgoprudny, Russia \\
    Federal Research Center “Computer Science and Control” \\
    Russian Academy of Sciences\\
    119333 Moscow, Russia\\
    AIRI\\
    123112 Moscow, Russia\\
	\texttt{panov@airi.net} \\
}

\renewcommand{\shorttitle}{Maneuver Decision-Making with Trajectory Streams Prediction}

\begin{document}
\maketitle

\begin{abstract}
Decision-making, motion planning, and trajectory prediction are crucial in autonomous driving systems. By accurately forecasting the movements of other road users, the decision-making capabilities of the autonomous system can be enhanced, making it more effective in responding to dynamic and unpredictable environments and more adaptive to diverse road scenarios. This paper presents the FFStreams++ approach for decision-making and motion planning of different maneuvers, including unprotected left turn, overtaking, and keep-lane. FFStreams++ is a combination of sampling-based and search-based approaches, where iteratively new sampled trajectories for different maneuvers are generated and optimized, and afterward, a heuristic search planner is called, searching for an optimal plan. We model the autonomous diving system in the Planning Domain Definition Language (PDDL) and search for the optimal plan using a heuristic Fast-Forward planner. In this approach, the initial state of the problem is modified iteratively through streams, which will generate maneuver-specific trajectory candidates, increasing the iterating level until an optimal plan is found. FFStreams++ integrates a query-connected network model for predicting possible future trajectories for each surrounding obstacle along with their probabilities. The proposed approach was tested on the CommonRoad simulation framework. We use a collection of randomly generated driving scenarios for overtaking and unprotected left turns at intersections to evaluate the FFStreams++ planner. The test results confirmed that the proposed approach can effectively execute various maneuvers to ensure safety and reduce the risk of collisions with nearby traffic agents.
\end{abstract}


\section{Introduction}
Autonomous driving systems are leading the charge in technological advancement, with the potential to revolutionize the future of transportation by allowing vehicles to drive and function on their own. These systems rely on several critical components, each essential for maintaining safe and efficient autonomous operation.

The key components of an autonomous driving system \cite{teng2023motion} are localization, perception, planning, and control. They work in concert to enable vehicles to navigate and operate autonomously in complex environments. By using and implementing advanced sensors, algorithms, and mapping technologies, autonomous driving systems hold the promise of safer, more efficient, and more accessible transportation for society as a whole \cite{yurtsever2020survey}.

Planning encompasses the process of generating a safe and efficient trajectory for the autonomous vehicle to follow \cite{10552884,gonzalez2019towards,palatti2021planning}. Having access to the HD map and leveraging data from the perception and localization modules and prediction model, planning algorithms analyze the surrounding environment, anticipate future scenarios relying on predictions of the future movements of other vehicles, pedestrians, and objects in the environment, and determine the optimal course of action \cite{yurtsever2020survey}. The action includes decisions on various maneuvers, such as yielding for or following a leading vehicle, lane keeping, lane changing, merging, overtaking, and turning at intersections.

\begin{figure}
	\centering
	\includegraphics[width=0.3\linewidth]{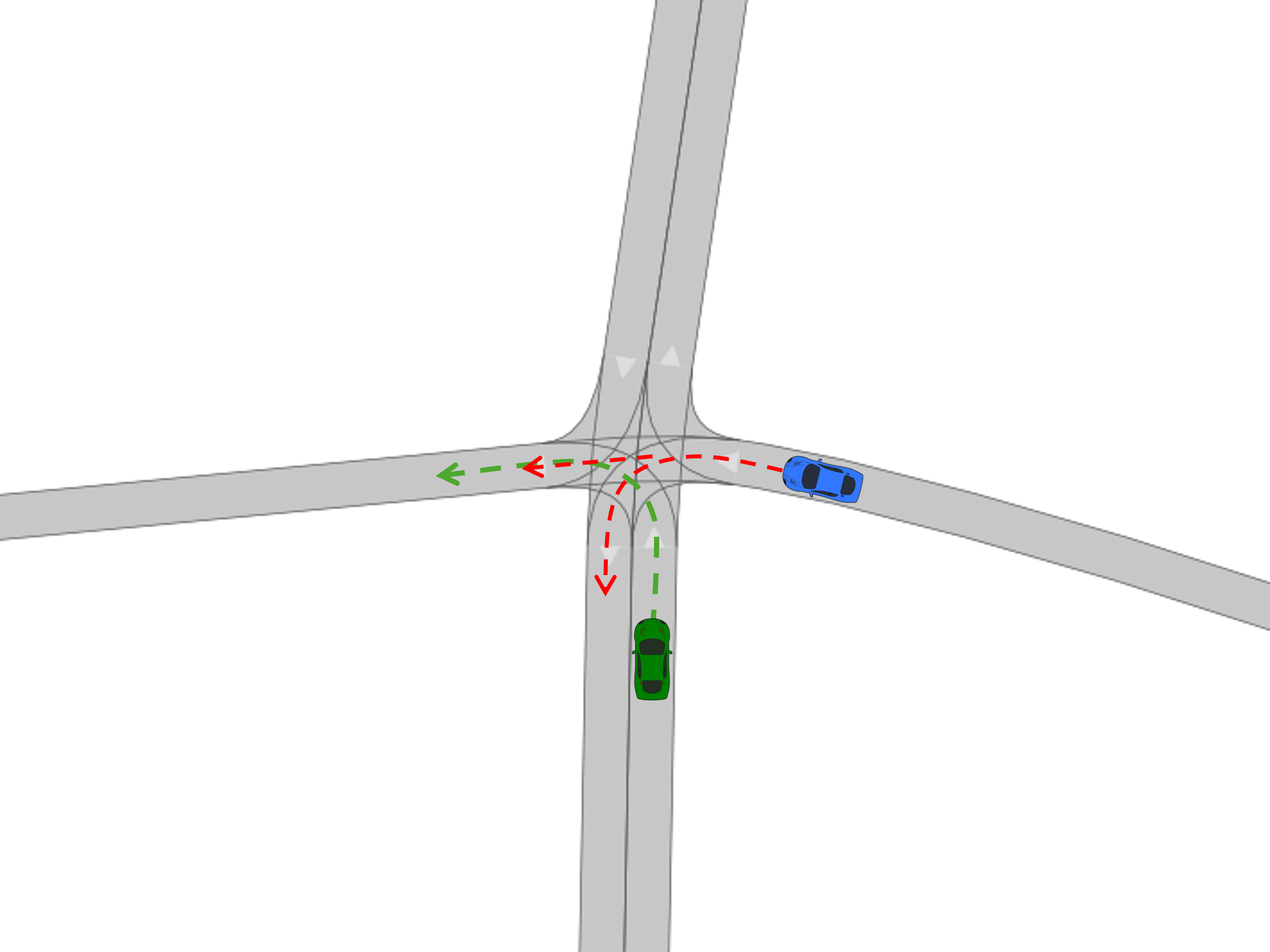}
	\caption{A critical unprotected left-turn maneuver at an unsignalized intersection.}
	\label{fig_intersection}
\end{figure}

\begin{figure}
	\centering
	\includegraphics[width=0.3\linewidth]{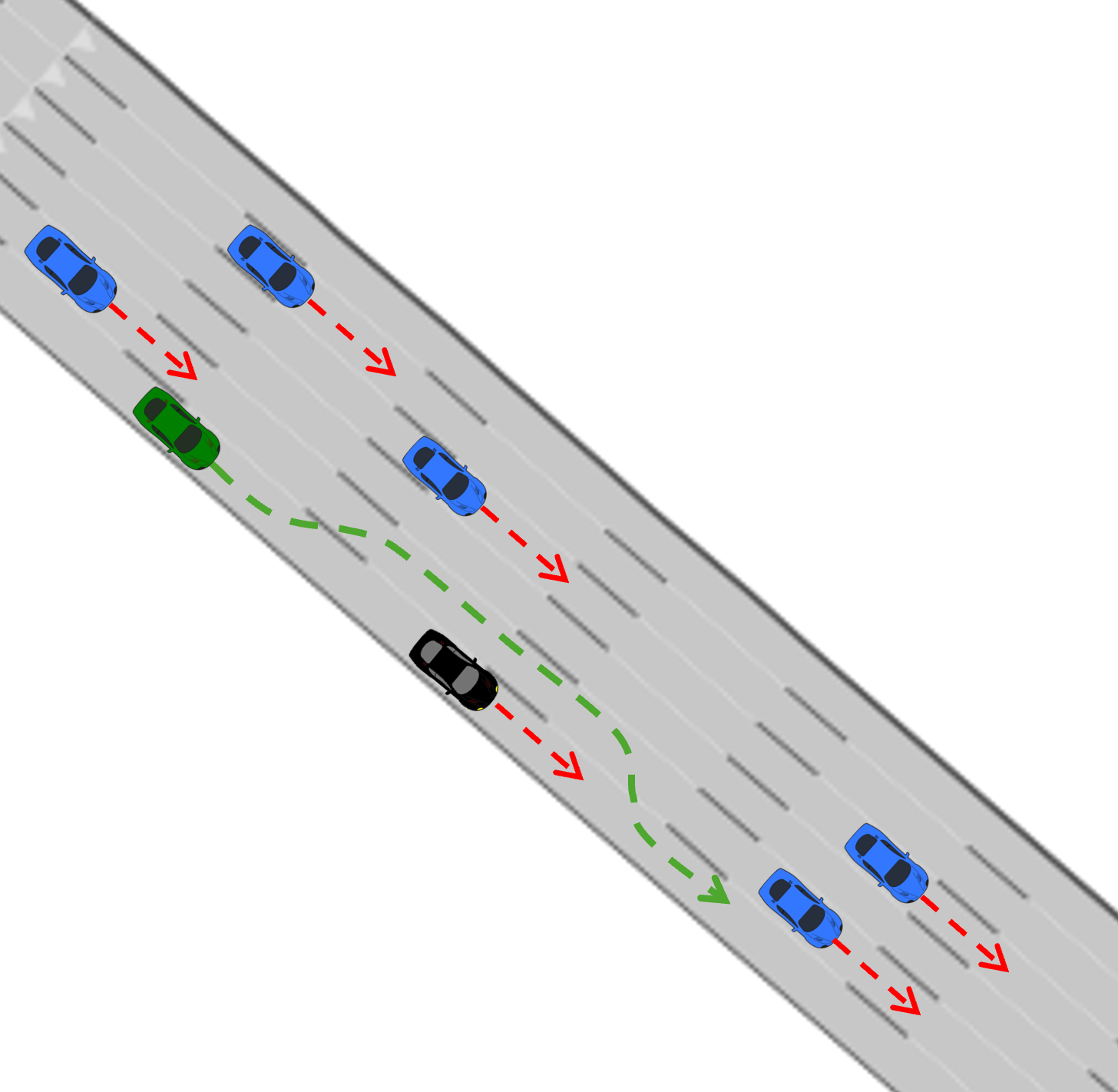}
	\caption{A critical overtaking maneuver in a highway scenario.}
	\label{fig_overtaking}
\end{figure}

A small error in decision-making, trajectory planning, and trajectory prediction can lead to potentially dangerous situations, including collisions and unsafe maneuvers. Figure \ref{fig_intersection} illustrates a critical unprotected left-turn maneuver at an unsignalized intersection where the oncoming vehicle's intention is unclear and the maneuver encounters high risk. Figure \ref{fig_overtaking} illustrates a critical overtaking maneuver when overtaking is desired when the front vehicle is moving at low speed and other surrounding vehicles are moving at high speed. 

By ensuring the accuracy of trajectory prediction models, planners can make safer decisions and reduce the risk of accidents \cite{garrido2022review}. Therefore, developing and testing planning algorithms with accurate trajectory prediction models is crucial for safe, efficient, and user-friendly autonomous driving systems. These models improve safety by anticipating the movements of surrounding objects, improve efficiency by optimizing trajectories, ensure robustness in diverse conditions, and enhance the user experience by providing smoother driving behavior.

Trajectory prediction methods can be classified \cite{bharilya2024machine} into conventional methods, deep learning-based methods, and reinforcement learning-based (RL) methods. Conventional methods encompass various approaches: Probabilistic Models estimate future trajectory probability distributions using Bayesian filters like Kalman and particle filters to handle prediction uncertainties \cite{li2019dynamic}; physics-based Models relying on the laws of physics and kinematics principles \cite{xie2017vehicle}. Learning-based approaches \cite{deo2018convolutional, jiang2023motion,cheng2023forecast,shi2022motion} utilize machine learning and deep learning on extensive driving datasets to predict trajectories, with recurrent neural networks (RNNs), convolutional neural networks (CNNs), and hybrid models showing promise. RL methods learn optimal policies for predicting future trajectories \cite{10003648}.

Conventional trajectory prediction methods are efficient and easy to implement but struggle with complex interactions and uncertainties. Advanced machine learning approaches, such as deep learning and RL, improve accuracy and robustness. Deep learning captures complex patterns and diverse scenarios, but it requires large labeled datasets and computational power, with challenges in model interpretability. RL methods learn from data and environmental interactions for accurate predictions but must balance algorithm complexity, training data availability, and generalization to various real-world scenarios.

Decision-making methods are classified into classical methods and leaning-based methods \cite{liu2021decision}. Classical decision-making methods include rule-based, optimization, and probabilistic approaches. Rule-based methods \cite{buehler2009junior,ziegler2014making,jamal2021adaptive} are interpretable and easy to implement, but they struggle with complex and dynamic conditions. Optimization methods \cite{bey2019optimization,artunedo2019decision} model interactions well but often fail in real-world scenarios due to their "optimal strategy" assumption. Probabilistic methods \cite{dong2017intention,isele2019interactive} combine well with others but have low efficiency in complex environments. Learning-based methods, such as statistical, deep learning, and reinforcement learning, offer varying advantages and drawbacks. Statistical methods are versatile but require extensive data and have low accuracy. Deep learning provides high accuracy and fully utilizes environmental data, but it needs large datasets and computational power, with interpretability challenges \cite{schwarting2018planning}.  Reinforcement learning handles uncertainty and dynamics well, but it faces several significant challenges \cite{dulac2019challenges,kiran2021deep}. These challenges include validating the performance of RL-based systems, bridging the simulation-reality gap, achieving sample efficiency, designing effective reward functions, and integrating safety into decision-making processes for autonomous agents, which can also lead to stability and overfitting issues.

\begin{figure}
	\centering
	\includegraphics[width=0.5\linewidth]{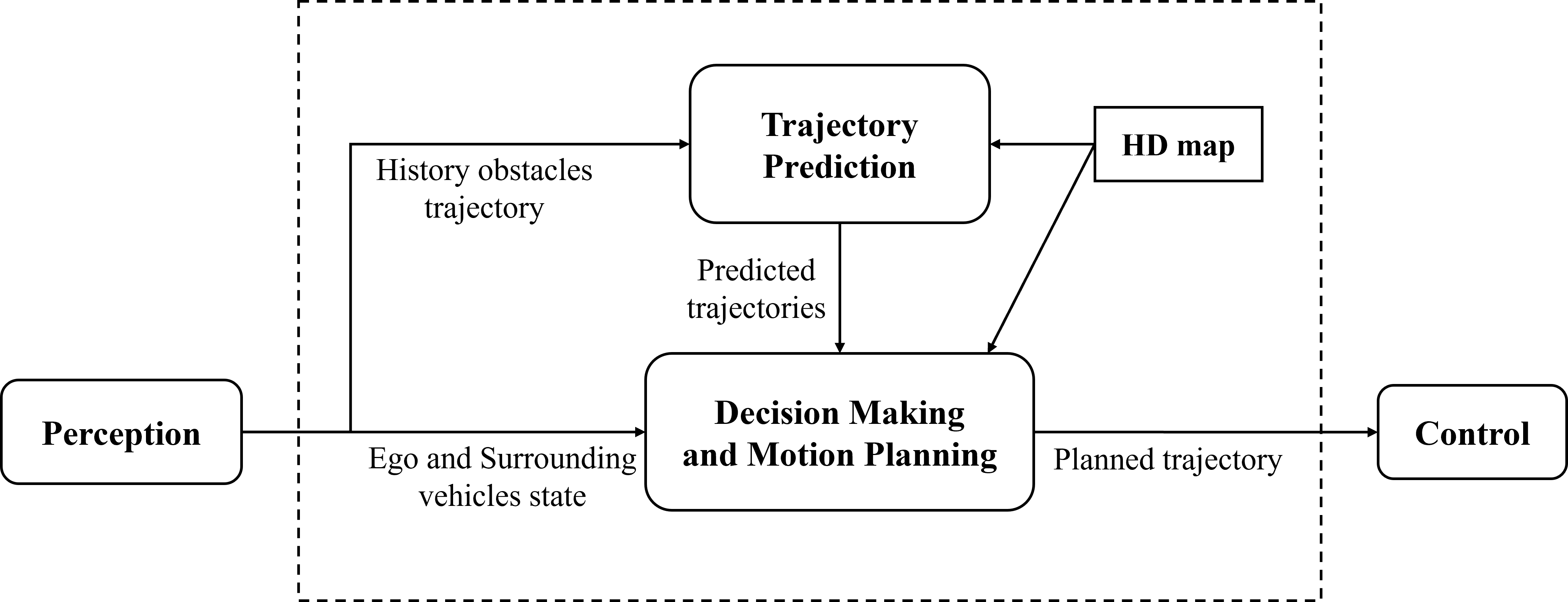}
	\caption{The schematic diagram of the proposed autonomous driving system is presented. It incorporates high-definition map information alongside the ego vehicle's current state and the present and past states of obstacles, as input from the perception module. The decision-making and motion planning framework, utilizing trajectory prediction, formulates the optimal trajectory, providing acceleration data and headings as outputs. These outputs serve as inputs to the control module within the autonomous driving system.}
	\label{fig_scheme}
\end{figure}

This paper introduces a new method for making decisions regarding vehicle maneuvers and planning motion, integrating trajectory prediction. The proposed method is designed for incorporation into an autonomous driving system. By utilizing data from localization and perception modules along with pre-existing high-definition (HD) map information of the environment, this approach determines optimal decisions and trajectories, including desired accelerations and headings for the ego vehicle. These outputs are then provided to the control module to generate the throttle and brake commands, enabling the ego vehicle to follow the planned trajectory. Figure~\ref{fig_scheme} illustrates the scheme of the proposed autonomous driving system. 

In previous work \cite{10552884}, the FFStreams framework was presented, addressing the problem of motion planning and decision-making for autonomous driving vehicles as an Integrated Task and Motion Planning problem. The FFStreams planner plans lane-keeping, overtaking, and lane-changing maneuvers and their associated trajectories by optimizing the jerk values while adhering to acceleration and curvature constraints to ensure a comfortable maneuver. In this paper, the FFStreams++ framework is introduced for planning unprotected left turns, overtaking, and lane-keeping, overcoming the challenges of FFStreams and integrating a trajectory prediction model for more realistic planning.

Due to its interoperability, expressiveness, and standardization, we represent the planning domain of autonomous driving in the Planning Domain Definition Language (PDDL). We integrate Streams into PDDL, introduced by \cite{garrett2020pddlstream}, where a Stream is a conditional function of an input set of object arguments. This function can modify the planning PDDL problem by generating an output tuple of new objects. Iteratively, we sample maneuver trajectories to update the initial PDDL problem's state using Streams and search for an optimal plan by heuristic Fast-Forward search. The FastForward (FF) heuristic search algorithm, developed by Jörg Hoffmann \cite{hoffmann2001ff}, is known for its effectiveness in solving PDDL due to its efficiency, scalability, and accuracy. The FF heuristic quickly estimates the cost of reaching the goal from a given state by ignoring delete effects, resulting in a significant speedup in finding solutions compared to more traditional search methods.

While the previously mentioned planning methods and planners have been successfully tested on a narrow range of maneuvers and with a simple physics-based prediction Model, we introduce a framework to plan a wide range of maneuvers, including lane-keeping (yield, speed follow), overtaking a front obstacle, passing intersection and unprotected left-turn at the intersection. In addition, we integrate into our framework a highly accurate prediction model Query-Centric network (QCNet) \cite{zhou2023query} trained on real-life scenarios. This work focuses on planning different maneuvers where two primary criteria guide our planning process: ensuring a safe trajectory and generating a comfortable maneuver by optimizing jerk values with acceleration and curvature constraints. This paper makes the following contributions:
\begin{itemize}
\item Introducing the FFStreams++ framework, which integrates decision-making and motion planning in the dynamic environment of autonomous driving, using maneuver-specific streams that optimize maneuver trajectory on jerk for planning comfortable maneuvers. 
\item Integration of prediction Query-Centric network (QCNet) model into the FFStreams++ framework for more reliable planning closer to reality. 
\item Validation of our approach through a set of experiments for overtaking and unprotected left turn maneuvers on real-life scenarios from CommonRoad benchmark \cite{commonroad}. The results demonstrate the effectiveness of the proposed method and the higher performance over a search-based planner.
\end{itemize}
The remainder of the paper is structured as follows: Section II discusses recent related work in trajectory prediction, decision-making, and motion planning. Section III provides background on automated planning in PDDL and streams. Section IV describes the methods, including the used benchmark, our trajectory prediction model, and the proposed FFStreams++ framework. Section V presents experiments and results, and Section VI provides an analysis of the experimental results. Section VII concludes with a discussion of future work.

\section{Related Work}

\subsection{Trajectory Prediction}

Deep learning-based approaches have recently gained considerable attention for trajectory prediction in autonomous vehicles. Artificial neural networks have proven highly efficient in trajectory prediction with the increased availability of extensive labeled training data for driving scenarios and enhanced computational resources for training and inference. These networks excel at identifying complex patterns and relationships from large datasets.

Extensive research \cite{shi2022motion,jiang2023motion,zhou2023query} has focused on improving the model architectures. Trajectory prediction models utilize a range of neural architectures \cite{bharilya2024machine}, including Recurrent Neural Networks (RNNs) such as Long Short-Term Memory Networks (LSTMs) and Gated Recurrent Units (GRUs), as well as Convolutional Neural Networks (CNNs), Graph Neural Networks (GNNs), Generative Adversarial Networks (GANs) and Transformer Networks, to forecast object movements. These models are adept at capturing both temporal and spatial dependencies in trajectory data, allowing them to accurately predict future trajectories. They are specifically designed to adapt to the complexity of the environment and the quality of the available sensor data.

Several researchers have proposed models that combine RNN and CNN architectures to manage temporal and spatial information for trajectory prediction. In their research \cite{deo2018convolutional}, the authors introduce an LSTM encoder-decoder model augmented with convolutional social pooling. This approach addresses the interactions between vehicles, using convolutional social pooling layers to learn interaction-related parameters more effectively than traditional fully connected layers. When evaluated with the publicly available NGSIM US-101 and I-80 datasets, the model performed better than a fully connected network regarding root mean squared prediction error and the negative log-likelihood of true future trajectories. The study also provides a qualitative analysis of the predicted distributions for various traffic scenarios. However, the model's reliance solely on vehicle tracks is highlighted as a limitation, as it misses out on incorporating map-based cues.

The authors of the study \cite{shi2022motion} have proposed a novel method based solely on transformer-based architecture that integrates map features with agent-specific features to predict the trajectory of a targeted agent. They employed an agent-centric, vectorized representation, converting all inputs into the local coordinates of each individual target. Although this approach demonstrates strong performance in predicting the trajectory of a single agent, it exhibits limited scalability and is challenging to implement when the number of agents in the driving environment increases—a common scenario in autonomous driving, where multiple agents are typically present.

In another study \cite{jiang2023motion}, researchers integrated Transformer encoders with CNNs to develop a motion query-based trajectory prediction model. This approach utilizes a Transformer encoder based on motion queries to capture temporal dependencies from historical trajectory sequences. Additionally, the model includes a social interaction module that incorporates convolutional networks and pooling operations to extract semantic information from the road and encode neighboring vehicle trajectories into a spatial social tensor. Evaluation of the NGSIM US-101 and I-80 datasets demonstrated enhanced prediction accuracy compared to previous methods. However, the model's performance is highly dependent on the quality of the spatial social tensor, with inaccuracies potentially degrading overall prediction accuracy. Additionally, in densely congested traffic scenarios, the complexity of processing dense spatial interactions may challenge the model’s efficiency and real-time performance.

Instead of relying on goal estimates, the authors of the study \cite{wang2023novel} used trajectory proposals as reference points in their approach. They introduced an agent-centric representation that employed agent characteristics in polar coordinates, which were then converted into Fourier features. While this method eliminated the need to recalculate agent-centric features at each time step, it is limited to a short prediction horizon. Specifically, the proposed method predicts the trajectories of surrounding agents for only three seconds, which is considered a short-term prediction. This limitation affects the reliability of the decision-making model that depends on it, as it may overlook crucial information about the future movements of the agents.

In their study \cite{zhou2023query}, the researchers introduced a Query-Centric Trajectory Prediction approach. Initially, they employed anchor-free queries to iteratively generate trajectory proposals. Subsequently, a refinement module used these proposals as anchors and applied anchor-based queries to further enhance the trajectories. While QCNet may face limitations in scenarios where the generated anchors do not accurately represent the ground truth, and its effectiveness could be challenged in dense, highly dynamic traffic scenes, it has nonetheless demonstrated superior predictive performance. Notably, QCNet excelled on the Argoverse 2 Motion Forecasting Benchmark, recognized for its extensive dataset, diverse scenarios, and rigorous evaluation metrics in the field of autonomous driving. Inspired by their approach, we adopt a similar structure in our method, ensuring compatibility with the expertise of the pre-trained neural network used in our evaluation.

\subsection{Decision-Making and Motion planning of maneuvers}

Decision-making strategies \cite{liu2021decision} in autonomous driving include Rule-based approaches (Finite State Machines, behavior trees, decision trees), probabilistic methods like Markov Decision Processes and variations, and learning-based approaches. Rule-based methods provide a clear understanding with deterministic outcomes prioritizing safety, yet they struggle with adapting to diverse and complex scenarios, posing challenges for maintenance.

Probabilistic-based approaches excel in adapting to diverse and uncertain driving conditions by evaluating the outcomes probabilistically. They effectively manage complexity and uncertain inputs, leveraging data and experiences to enhance decision-making through machine learning. However, they heavily rely on high-quality training data, are susceptible to biases and inaccuracies that can affect reliability and performance, and have a risk of overfitting.

Learning-based approaches offer adaptability and real-time robustness in complex environments, outperforming rule-based systems. They require extensive training data, are prone to overfitting, and lack transparency in decision-making processes.

In their research \cite{palatti2021planning}, the authors improved reactive planning by introducing a method for autonomous overtaking that integrates safety measures, including the ability to cancel the overtaking maneuver if safety is jeopardized. Their strategy divides the task into behavior and trajectory planning stages, employing a finite state machine (FSM) based on heuristic rules for maneuver planning followed by a Model Predictive Control (MPC)--based trajectory planner to generate collision-free paths. However, utilizing a simple FSM might be insufficient for complex situations. Designing an FSM to accommodate all potentialities of a complex environment is impractical, and simpler FSMs or rule sets may either sacrifice performance due to excessive caution or compromise safety due to over-optimism.

The lane change decision-making was structured as a Partially Observable Markov Decision Process (POMDP) in the paper \cite{gonzalez2019towards}, integrating a behavioral model derived from driving data demonstrations. Monte Carlo simulations were used via the history tree to estimate the action node values. However, this method faces limitations as it does not adhere to real-time requirements crucial for autonomous driving systems. These limitations stem from costly belief updates and the need to reconstruct the history tree for each planning step due to the continuous observation space.

In their research \cite{li2022combining}, authors employed an RL agent integrated with a deep neural network for decision-making and trajectory planning in an automated lane change system. The RL agent is trained to identify the target lane and decide to stay in the current lane or switch to the target lane. The agent optimizes lane changes using a deep neural network to minimize overall travel times. A unified trajectory planning model also generates a reference path and velocity profile when a lane-keeping or changing decision is made. However, the method may lack adaptation to real-world road scenarios that may include mixed straight and curved sections, indicating the need for further research to enhance its effectiveness and generalization.

In another study \cite{schmidt2019maneuver}, authors presented a framework for lane change behavior planning emphasizing a convex optimization-based approach. Initially, a spatial-temporal depiction of the traffic scenario (the maneuver) is identified by polygon clipping based on trajectory predictions of obstacle vehicles. Subsequently, safety constraints are integrated into the trajectory optimization phase via convex quadratic programming, specifically focusing on Time-To-Collisions and Time Gaps.

In the research \cite{yuan2024evolutionary}, authors developed a framework for decision-making and trajectory planning to perform lane-changing maneuvers on a highway. Their framework used a data-driven decision-making module based on deep reinforcement learning (DRL) to plan the driving behavior. Then, they employed a model predictive control (MPC), which outputs acceleration and steering commands to execute longitudinal and lateral motion planning tasks. For the online evolution of the DRL decision-making module, the authors used a combination of a predictive safe-driving envelope model and a rational scheme.

In a recent study \cite{10552884}, the FFStreams framework was introduced to tackle the challenges of motion planning and decision-making in autonomous driving vehicles using an Integrated Task and Motion Planning approach. The FFStreams planner is designed to plan lane-keeping, overtaking, and lane-changing maneuvers along with their respective trajectories. This is achieved by formulating maneuver planning as a PDDL problem. Initially, a base PDDL problem is defined, which is then dynamically modified using configuration and trajectory streams. These streams introduce new predicates representing candidate maneuver trajectories. The framework employs FastForward search to iteratively search for the optimal plans. The trajectories generated by the streams are optimized on jerk in the Frenet coordinate system while ensuring adherence to acceleration and curvature constraints to facilitate smooth maneuvers. The proposed framework integrates a simplified prediction module that assumes all dynamic obstacles maintain a constant velocity. Although the framework achieves real-time performance and rapidly adapts to changes in obstacle velocities by replanning after each new observation, it may encounter challenges in highly dynamic environments.

Previous research in maneuver planning has typically focused on individual maneuvers under simplified conditions and specific scenarios. In contrast, our work introduces FFStreams++, an enhanced framework for maneuver planning that integrates trajectory prediction into FFStreams, enabling it to handle a variety of maneuvers in diverse scenarios, including highway and intersection settings. This extension incorporates a trajectory prediction model for surrounding obstacles, enhancing the planner's capability to optimize trajectories and make decisions across different maneuvers such as lane-keeping, yielding at intersections, unprotected left turns, lane changing, and overtaking.

\begin{figure}
	\centering
	\includegraphics[width=\textwidth]{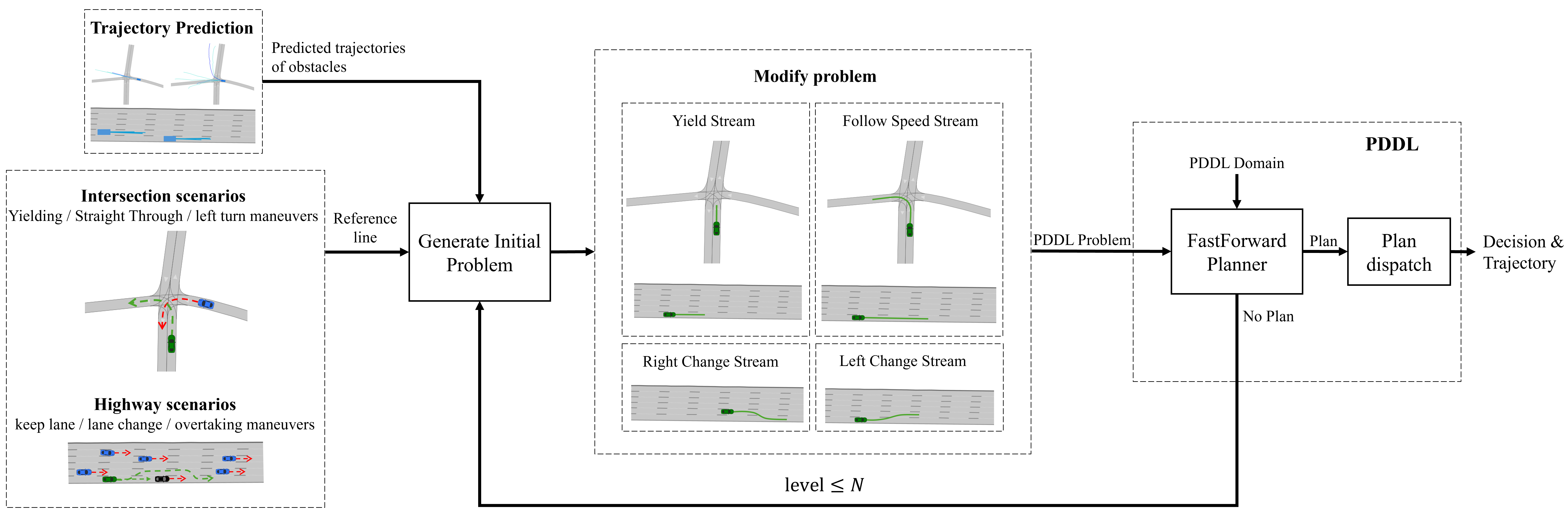}
	\caption{Scheme of the proposed FFStreams++ framework for integrated decision-making and motion planning with trajectory prediction in autonomous driving within dynamic environments. The system combines trajectory prediction with a target reference line derived from intersection and highway scenarios to generate an initial PDDL problem. This problem is iteratively modified by considering various trajectory streams (Yield, Follow Speed, Right Change, and Left Change Streams) until an optimal plan is achieved. The problem is formulated in PDDL2.1, solved using a FastForward Planner, and the resulting optimal plan is dispatched to guide the vehicle's decision-making and trajectory.}
	\label{all_schema}
\end{figure}

\section{Background}

Planning Domain Definition Language (PDDL) is a formal language utilized to describe planning problems in artificial intelligence. To effectively model a planning problem in PDDL, two primary components are required: the domain and the problem. The domain $(Dom)$ defines the general actions and objects involved, while the problem $(Prob)$ specifies the initial state and desired goals for a particular instance of the domain. PDDL2.1 \cite{fox2003pddl2} introduced numeric fluents, and plan metrics, enhancing the language by allowing the integration of optimization metrics within planning problems.

The planning domain $Dom = \langle T, P, F, A\rangle$ consists of: \textit{Types T} - a set of types used to classify objects in the domain, \textit{Predicates P} - a set of predicates, \( P = \{p_1, p_2, \ldots, p_p\} \), where each predicate \( p \) can be applied to a specific sequence of objects $\{o_1, o_2, \ldots, o_o\}$ to form literals, \textit{Functions F} - a set of functions(variables), \( F = \{f_1, f_2, \ldots, f_f\} \), where each fuction \( f \) can be applied to a specific sequence of objects assigning a value to them, \textit{Actions A} - a set of actions, \( A = \{a_1, a_2, \ldots, a_j\} \), where each action \( a \) is defined by a tuple of object arguments $\bar{o} = \langle o_{1}, o_{2},..., o_{r}\rangle$ and a set of preconditions $pre(a)$ of $\bar{o}$, which are positive literals $pre^{+}(a)$ and negative literals $pre^{-}(a)$ that must hold for the action to apply, and a set of effects $eff(a)$ on $\bar{o}$, which are positive literals $eff^{+}(a)$ and negative literals $eff^{-}(a)$ that are the result of applying the action.

The action $a$ is applicable at the state $I$ if 
\begin{equation}
    \left(\right.pre\left.^{+}(a(\bar{o})) \subseteq I\right) \wedge\left(\right.pre\left.^{-}(a(\bar{o})) \cap I=\emptyset\right)
\end{equation}
The resulting state $I'$ after applying action $a$ in a state $I$:
\begin{equation}
    I' =\left(I \backslash eff^{-}(a(\bar{o}))\right) \cup eff^{+}(a(\bar{o}))
\end{equation}

The planning problem $Prob = \langle O, I_{0}, G\rangle$ consists of \textit{Objects $O$} - a set of objects that are instances of the defined types, \textit{Initial State $I_{0}$} - a set of positive literals expressing the initial state, \textit{Goal State $G$} - a set of both positive and negative literals expressing the goal state. A plan $\pi=\left[a_{1}\left(\bar{o}_{1}\right), \ldots, a_{k}\left(\bar{o}_{k}\right)\right]$ is a finite sequence of $k$ action instances such that each $a_{i}\left(\bar{o}_{i}\right)$ is applicable in the state $I_{i-1}$ leading to the state $I_{i}$. The goal state $G$ is satisfied after applying the entire sequence of $\pi$.

A stream $s(\bar{o})$ is a conditional function of an \emph{input} set of object arguments $\bar{o} = \langle o_{1},o_{2},...,o_{m}\rangle$. This function can modify the planning problem $Prob$ by generating a \emph{output} tuple of new objects $\bar{r} = \langle r_{1},r_{2},...,r_{l}\rangle$, and a set of certified facts associated with them, where $s.cert=\{p \mid \forall \bar{o} \in \bar{O}, \forall \bar{r} \in s(\bar{o}) . p(\bar{o}+\bar{r})\}$. The stream can yield None if generating new objects is impossible. The stream can be applied on input parameters $\bar{o}$ only if a set of positive literals $p$ related to them exists in the domain, where $s.dom=\{p \mid \forall \bar{o} \in \bar{o} \cdot p(\bar{o})\}$. A stream can modify the initial state $I_0$ in the problem $Prob$.

When streams are integrated into automated planning in PDDL, the process is iteratively run at various planning levels. Initially, the applicable streams are used to modify the initial state. Subsequently, the PDDL planner performs a heuristic search to find an optimal plan. If no plan is found, the planning level is increased, and the process repeats. This iterative procedure continues until either an optimal plan is discovered or the maximum planning level is reached.

One of the efficient planners for solving a PDDL problem is the FastForward (FF) planner. The FF planner employs heuristic search within the planning space, systematically exploring applicable actions and new states until the goal state is achieved. It utilizes the heuristic $h^{FF}$, which is based on the length of the relaxed plan. A relaxed plan is a simplified version of the problem that ignores the negative effects $eff^{-}(a(\bar{o}))$ of the actions. The $h^{FF}$ heuristic is admissible, returning infinity if no relaxed plan exists and otherwise providing the length of the relaxed plan, indicated by the number of action layers in the relaxed planning graph.

\section{Methods}

We propose an FFStreams++ framework to solve the integrated decision-making and motion-planning problem (Figure \ref{all_schema}). We integrate a trajectory prediction Query-Centric network (QCNet) into FFStreams++. In the next subsections, we explain each part in detail.

\subsection{QCNet Trajectory Prediction Model}
A pre-trained Query-Centric network (QCNet) \cite{zhou2023query} was used for predicting the trajectories of surrounding vehicles. The QCNet was trained on the Argoverse 2 motion forecasting benchmark. The set of scenarios extracted from the CommonRoad benchmark is within the scope of the pre-trained neural network's expertise.

The network's input is map information and the states of obstacles at the last $T$ time steps. The state includes obstacles' 2D positions and headings. Map information includes polygons of the high-definition map (e.g., lanes and crosswalks), where each map polygon is annotated with sampled points and semantic attributes (e.g., the user type of a lane).
The QCNet predicts each obstacle's future $K$ trajectories over a prediction horizon of $T'$ and provides a probability score for each trajectory. 

The QCNet was adjusted by converting the map information from Lanelet2 format to Argoverse HD map format and the agents' states.

Our prediction model was trained to predict six possible future trajectories for each surrounding obstacle and their probabilities. In sequence, we input the two predicted trajectories with the highest and second-highest probabilities for each obstacle into the FFStreams++ planner.

Choosing an autonomous driving algorithm's prediction and planning horizons involves balancing several factors to ensure optimal performance. When choosing the prediction horizon, factors such as traffic environment complexity (urban area, highway), vehicle speed, and model capability should be taken into consideration. Longer prediction horizons increase uncertainty and can reduce prediction reliability. Short horizons can cause valuable information to be lost about surrounding vehicles. When choosing a planning horizon, factors such as route complexity, computational resources, and response time should be taken into consideration. Longer planning horizons provide smoother and more comfortable rides but may be less responsive to sudden changes as they require more computational power and can increase latency.

\begin{figure}
	\centering
	\includegraphics[width=0.55\linewidth]{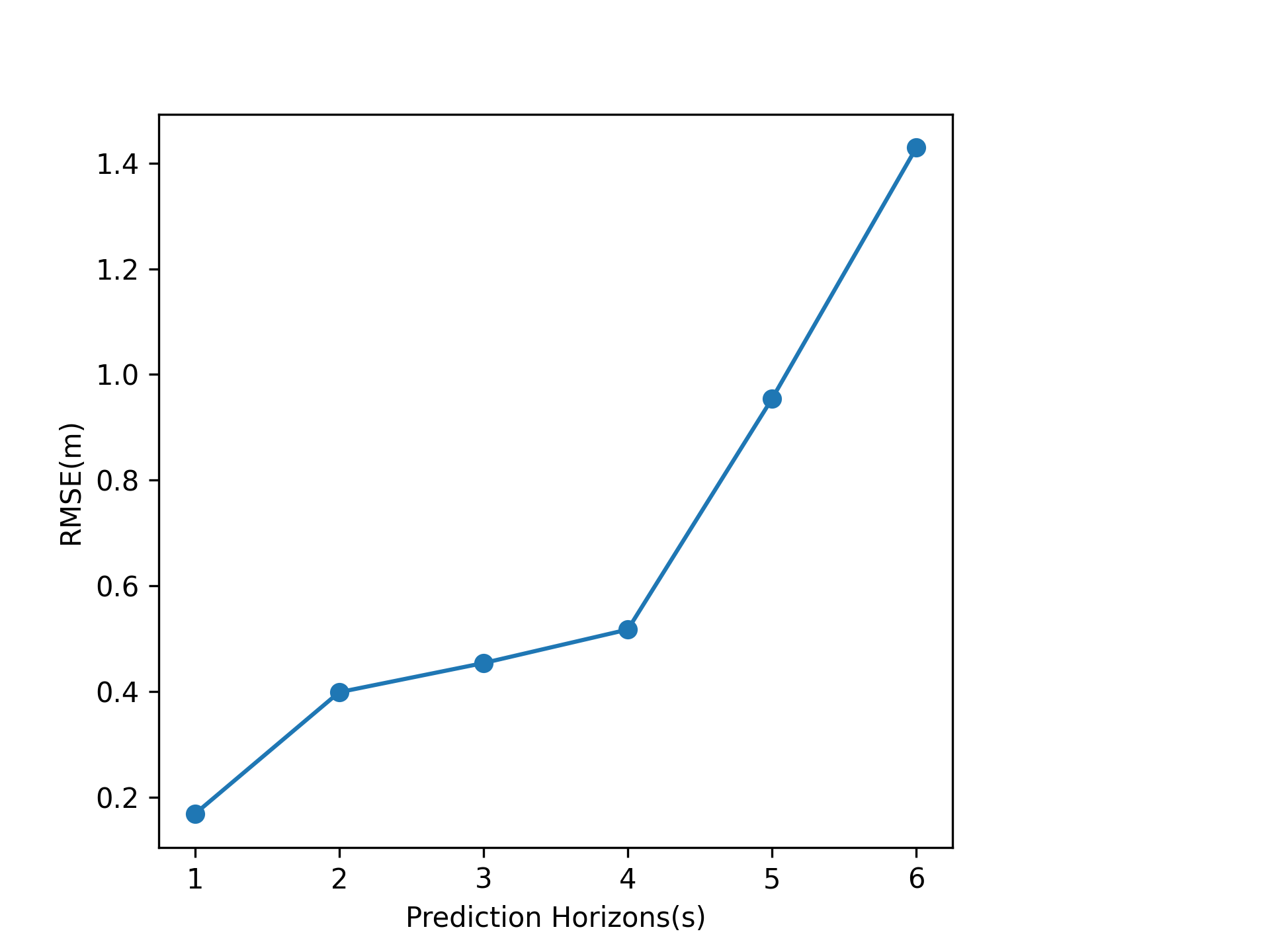}
	\caption{The RMSE of the prediction model against different prediction horizons.}
	\label{fig_prediction}
\end{figure}

We predict the future trajectories of obstacles on a prediction horizon of $5$ seconds, which was chosen after evaluating the Root Mean Squared Error (RMSE) metric on different prediction horizons (Figure \ref{fig_prediction}) on various highway CommonRoad scenarios. The RMSE measures the $L_2$ distance in meters between the ground-truth trajectory and the best of predicted trajectory, averaged over all future time steps.

We also chose the planning horizon to be $5$ seconds after examining the algorithm's runtime, rider comfort, and smoothness of trajectories.

\subsection{FFStreams++ Decision-Making and Motion Planning Framework}
\label{sec:framework}

The FFStreams++ framework is adapted to plan unprotected left turn maneuvers in addition to lane-keeping, lane-changing, and overtaking maneuvers and to consider more precise predicted trajectories of surrounding obstacles to enhance the planned behavior and motion, considering the obstacles' accelerations.

The FFStreams++ framework represents the planning domain and problem in PDDL language, edits the planning problem using streams, and searches for the optimal plan using FastForward heuristic search. Streams are used to generate an optimized trajectory on Frenet for each maneuver, along with associated predicates that will be added to the PDDL problem.

\subsubsection{PDDL Domain and FastForward Search}
We formulate the Decision-making and motion-planning problem as PDDL2.1 planning domain $(Dom)$ (representing predicates, their types, and possible actions) and PDDL2.1 planning problem $(Prob)$ (representing objects, initial state predicates, goal predicates, plan optimization metric). We use the extension PDDL2.1 for numeric planning to represent different predicates with numeric values, and for performing arithmetic operations in PDDL to calculate the distance to obstacles during collision checking action.

In the PDDL domain, we define two types of objects: car configuration $-conf$ and obstacle $-obstacles$. We define the set of predicates: \emph{(yield\_traj $?q_{1}$ $?q_{2}$)},\emph{(keep\_speed\_traj $?q_{1}$ $?q_{2}$)},\emph{(overtake\_traj $?q_{1}$ $?q_{2}$)},\emph{(left\_traj $?q_{1}$ $?q_{2}$)}, and \emph{(right\_traj $?q_{1}$ $?q_{2}$)} indicating the existence of a trajectory from configuration $q_{1}$ to configuration $q_{2}$ for yielding (decelerating), following speed (accelerating), overtaking, changing to left lane or changing to right lane, \emph{(next $?q_{1}$ $?q_{2}$ $?q_{end}$)} indicating a sequence of sub-configurations on a trajectory to the final configuration $q_{end}$, \emph{idle} stating that the collision checking process is idle to start a new check, \emph{(checking\_traj $?q_{1}$ $?q_{2}$ $?o$)} indicating that the trajectory between configurations $q_{1}$ and $q_{2}$ is under checking with obstacle $o$, \emph{(checked\_traj $?q_{1}$ $?q_{2}$ $?o$)} indicating that the trajectory between the two configurations is checked with obstacle $o$ and is collision-free, \emph{(ego\_at $?q$)} stating that the ego vehicle is currently at configuration $?q$, \emph{(on\_init\_lane)}, \emph{(on\_second\_lane)}  stating the ego's current lane.

The functions of the maneuver-planning domain include the following changing variables: \emph{(total\_cost)} - total cost, \emph{(curr\_time)} - current time, \emph{(time\_of\_traj $?q_{1}$ $?q_{2}$)} - the duration of following the trajectory, \emph{(at\_x $?q$)} \emph{(at\_y $?q$)} \emph{(at\_time $?q$)} - coordinates and the time step of a certain configuration, \emph{(obst\_at\_x $?o$ $?q$)} \emph{(obst\_at\_y $?o$ $?q$)} - coordinates of a certain obstacle at a certain ego's configuration, implicitly associating obstacles' coordinates with a specific time step. 

The maneuver domain consists of a set of actions $A$ for maneuvers: \textit{keep\_speed} action to keep the lane while accelerating, \textit{keep\_yield\_speed} action to keep the lane while decelerating, \textit{left\_change}/\textit{right\_change} actions to move to a neighbor lane, \textit{overtake} action to overtake a front obstacle if exists. We define the set of Actions $A$ as follows:
$$A = \{keep\_speed(\bar{o}),keep\_yield\_speed(\bar{o}),$$
$$left\_change(\bar{o}),right\_change(\bar{o}),overtake(\bar{o})\}.$$

The partial PDDL Domain is demonstrated below.

\begin{lstlisting}[float=!htb,language=PDDL,basicstyle=\fontsize{7.5}{7.5}\selectfont,belowskip=-0.8\baselineskip]
(:predicates  
    (next ?q1 ?q2 ?q_end - conf) (traj ?q1 ?q2 - conf) (idle) (is_first ?q - conf ?o - obstacles)
    (is_last ?q - conf ?o - obstacles) (ego_at ?q - conf) (checking_traj ?q1 ?q2 - conf ?o - obstacles)
    (checked_traj ?q1 ?q2 - conf ?o - obstacles) (moved_forward) (there_is_front_obs) (on_init_lane)
    (on_second_lane) (yield_traj ?q1 ?q2) (keep_speed_traj ?q1 ?q2) (overtake_traj ?q1 ?q2)
    (left_traj ?q1 ?q2) (right_traj ?q1 ?q2))
(:functions (cost)(curr_time) (time_of_traj ?q1 ?q2 - conf) (at_x ?q - conf)(at_y ?q - conf) 
    (at_time ?q - conf) (obst_at_x ?o - obstacles ?q - conf) (obst_at_y ?o - obstacles ?q - conf) 
)
(:action keep_speed
    :parameters (?q1 ?q2 - conf)
    :precondition (and 
         (ego_at ?q1) (traj ?q1 ?q2) (keep_speed_traj ?q1 ?q2) (on_init_lane) (idle)
         (forall (?o - obstacles)
            (checked_traj ?q1 ?q2 ?o))
    )
    :effect (and 
         (ego_at ?q2) (not (ego_at ?q1)) (increase (curr_time)  (time_of_traj ?q1 ?q2))
         (increase (cost)  5) (moved_forward)
    )
)
(:action keep_lane_yield
    :parameters (?q1 ?q2 - conf)
    :precondition (and  (ego_at ?q1) (traj ?q1 ?q2) (yield_traj ?q1 ?q2) (on_init_lane) (idle)
         (forall (?o - obstacles)
            (checked_traj ?q1 ?q2 ?o) )
    )
    :effect (and 
         (ego_at ?q2) (not (ego_at ?q1)) (increase (curr_time)  (time_of_traj ?q1 ?q2)) 
         (increase (cost)  10) (moved_forward)
    )
)
(:action left_change
    :parameters (?q1 ?q2 - conf)
    :precondition (and 
         (ego_at ?q1) (traj ?q1 ?q2) (left_traj ?q1 ?q2) (on_init_lane) (idle)
         (forall (?o - obstacles)
            (checked_traj ?q1 ?q2 ?o) )
    )
    :effect (and 
         (ego_at ?q2) (not (ego_at ?q1)) (not (on_init_lane)) (increase (cost)  1)
         (increase (curr_time)  (time_of_traj ?q1 ?q2)) (moved_forward) 
    )
)
\end{lstlisting}

After defining the planning domain and planning problem $(Dom,Prob)$, the FastForward heuristic planner searches for a plan with minimal cost. The FF heuristic $h^{FF}$ operates on a relaxed version of the planning problem where the delete effects of actions are ignored. This means that once a fact (predicate) becomes true, it remains true, simplifying the problem. The FF heuristic constructs a Relaxed Planning Graph (RPG), a layered graph where each layer represents a set of facts or actions that could be achieved or applied at that step without considering delete effects. Once the goal predicates are present in a fact layer, the algorithm traces back through the graph from the goals, selecting actions that achieve the goals. The sum of the costs of these actions provides the heuristic estimate. Guided by these heuristic estimates, FF uses a weighted A* search strategy to explore the state space and find a solution. The time complexities for Relaxed Planning Graph construction and heuristic calculation are both polynomials.

\subsubsection{Maneuver Streams with Frenet and Jerk Optimization}

We use a stream for each maneuver, following speed, yielding to a front obstacle or at the intersection, changing to the left neighbor lane, and changing to the right neighbor lane. The applicable streams will be called at each iteration of the FFStreams++ algorithm. As each stream has its conditions, not all of them are applicable. For instance, overtake\_trajectory stream is only applicable if there is a front obstacle. 

\begin{lstlisting}[float=!htb,language=PDDL,basicstyle=\fontsize{7.5}{7.5}\selectfont,belowskip=-0.8 \baselineskip]
(:stream yield_stream    
    :inputs (?q1)   
    :domain (and (at_x ?q1) (at_y ?q2) (at_time ?q1))
    :outputs (?q2 ?q_{1_2_1} ... ?q_{1_2_24})
    :certified 
        (and (yield_traj ?q1 ?q2)
        (time_of_traj ?q1 ?q2) 
        (at_x ?q2) (at_y ?q2) (at_time ?q2)
        (next ?q1 ?q1_2_1 ?q2) ... 
        (next ?q1_2_23 ?q1_2_24 ?q2)))
        
(:stream overtake_stream   
    :inputs (?q1)   
    :domain (and (at_x ?q1) (at_y ?q2) (at_time ?q1)
        (there_is_front_obs))
    :outputs (?q2 ?q_{1_2_1} ... ?q_{1_2_24})
    :certified 
        (and (overtake_traj ?q1 ?q2)
        (time_of_traj ?q1 ?q2) 
        (at_x ?q2) (at_y ?q2) (at_time ?q2)
        (next ?q1 ?q1_2_1 ?q2) ... 
        (next ?q1_2_23 ?q1_2_24 ?q2)))
\end{lstlisting}

For each maneuver stream, we define $v_{desired}$ a desired final speed of the trajectory according to the maneuver type. To generate a trajectory candidate for each maneuver, we use Frenet optimization for trajectory planning and integrate jerk optimization into the cost function to select an efficient and comfortable trajectory.

We define the reference path as the centerline of the desired lane and the initial state of the ego vehicle as $[x_0,y_0,\theta_0,v_0,a_0]$, where $(x_0,y_0),\theta,v_0,a_0$ are the initial Cartesian coordinates, heading, velocity and acceleration of the ego vehicle. We then convert the Cartesian coordinate system to the Frenet coordinate system, defining the state in the Frenet system as:
\begin{equation}
x(t)=[s(t),s'(t),s''(t),s'''(t),l(t),l'(t),l''(t),l'''(t)]
\end{equation}
where $s$: longitudinal position, $s'$: longitudinal velocity, $s''$: longitudinal acceleration, $s'''$: longitudinal jerk, $l$: lateral position, $l'$ : lateral velocity, $l''$ : lateral acceleration, and $l'''$: lateral jerk.

Knowing the initial Frenet state $[s_0,s'_0,s''_0,l_0,l'_0,l''_0]$, a set of lateral and longitudinal trajectories are generated using quintic polynomials for lateral movement and quartic polynomials for longitudinal movement. Trajectories that exceed the maximum acceleration $a_{max}$, the maximum speed $v_{max}$, or the maximum curvature $\kappa$ are excluded from the set. Afterward, For each trajectory $\tau$ the overall cost $J$ is evaluated, given as:
\begin{equation}
J(\tau) = J_{comfort}(\tau) + J_{efficiency}(\tau) + J_{lateral\_error}(\tau) + J_{speed\_error}(\tau),
\end{equation}
where $J_{comfort}$ is the cost related to passenger comfort, including jerk minimization to penalize trajectories with high jerk values, $J_{efficiency}$ is the cost related to travel time, $J_{lateral\_error}$ is the cost related to the lateral error at the final point and $J_{speed\_error}$ is the cost related to the longitudinal speed error at the final point. We define the previously mentioned costs as:
\begin{equation}
J_{comfort} =  \omega_j \int_0^T(s'''^2 + l'''^2)dt
\end{equation}
\begin{equation}
J_{efficiency} = \omega_t.T
\end{equation}
\begin{equation}
J_{lateral\_error} = \omega_{err}.(l(T) - l_{desired})^2 = \omega_{err}.e_l^2
\end{equation}
\begin{equation}
J_{speed\_error} = \omega_{err}.(s'(T) - s'_{desired})^2 = \omega_{err}.e_{s'}^2,
\end{equation}
where $\omega_j,\omega_t,\omega_{err}$ are weighting factors for jerk, travel time, and error costs, $T$ is travel time, $e_l$ is lateral error, $e_{s'}$ is the longitudinal speed error, $s'_{desired}$ is the desired speed at final point, and $l_{desired}$ is the desired lateral position at final point. As the goal is to follow the reference path $l_{desired}$ is set to zero. The parameters of optimization are stated in Table~\ref{parameterstable}.

The overall cost is:
\begin{equation}
J(\tau) = \omega_j \int_0^T(s'''^2 + l'''^2)dt + \omega_t.T + \omega_{err}.(e_l^2+e_{s'}^2).
\end{equation}
The optimal trajectory $\tau^*$ is the trajectory that minimizes the overall cost:
\begin{equation}
\tau^* = argmin_{\tau} J(\tau).
\end{equation}

\begin{table}[!t]
\caption{Frenet and Jerk Optimization Parameters\label{parameterstable}}
\centering
\begin{tabular}{lll}
\toprule
Parameter & Symbol & Value\\
\midrule
Jerk weight  &$\omega_j$  &  $0.1$\\
Travel time weight&$\omega_t$  &  $0.1$\\
Error weight&$\omega_{err}$  &  $1.0$\\
Planning horizon&$T$  &  $5.0$ $[s]$\\
Time step & $\Delta t$ & $0.2$ $[s]$\\
Maximum acceleration& $a_{max}$  & $2.0$ $[m/s^2]$\\
Maximum speed   &  $v_{max}$ & $57.6$ $[m/s]$ \\ 
Maximum Curvature  & $\kappa$ &$1$ $[1/m]$ \\
\bottomrule
\end{tabular}
\end{table}

\begin{figure*}[t!]
 \centering
 \includegraphics[width=\textwidth]{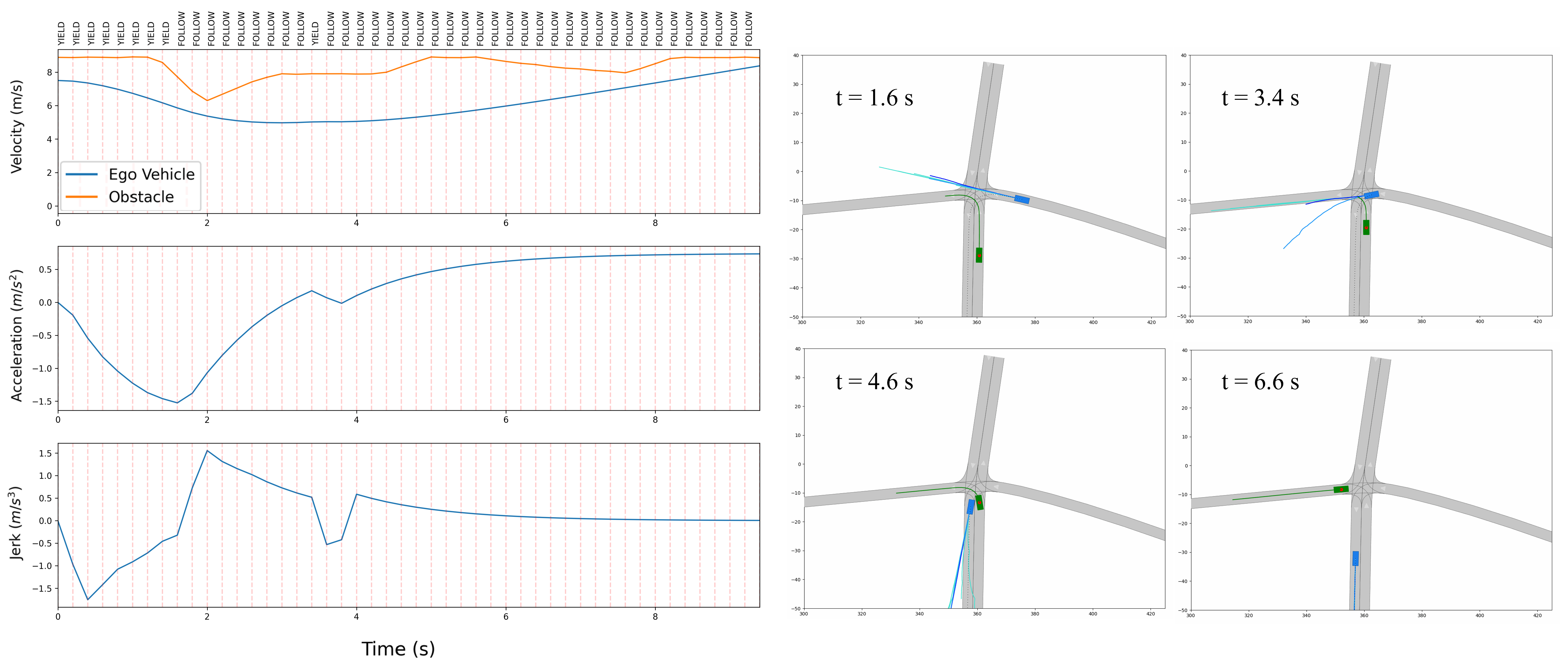}
 \caption{A successful experiment of unprotected left-turn maneuver planned by FFStreams++ planner and the velocity, acceleration, and jerk profile of the planned trajectories as well as the decision at each timestep.}
 \label{fig_profiles_left}
\end{figure*}

\begin{figure*}[t!]
 \centering
 \includegraphics[width=\textwidth]{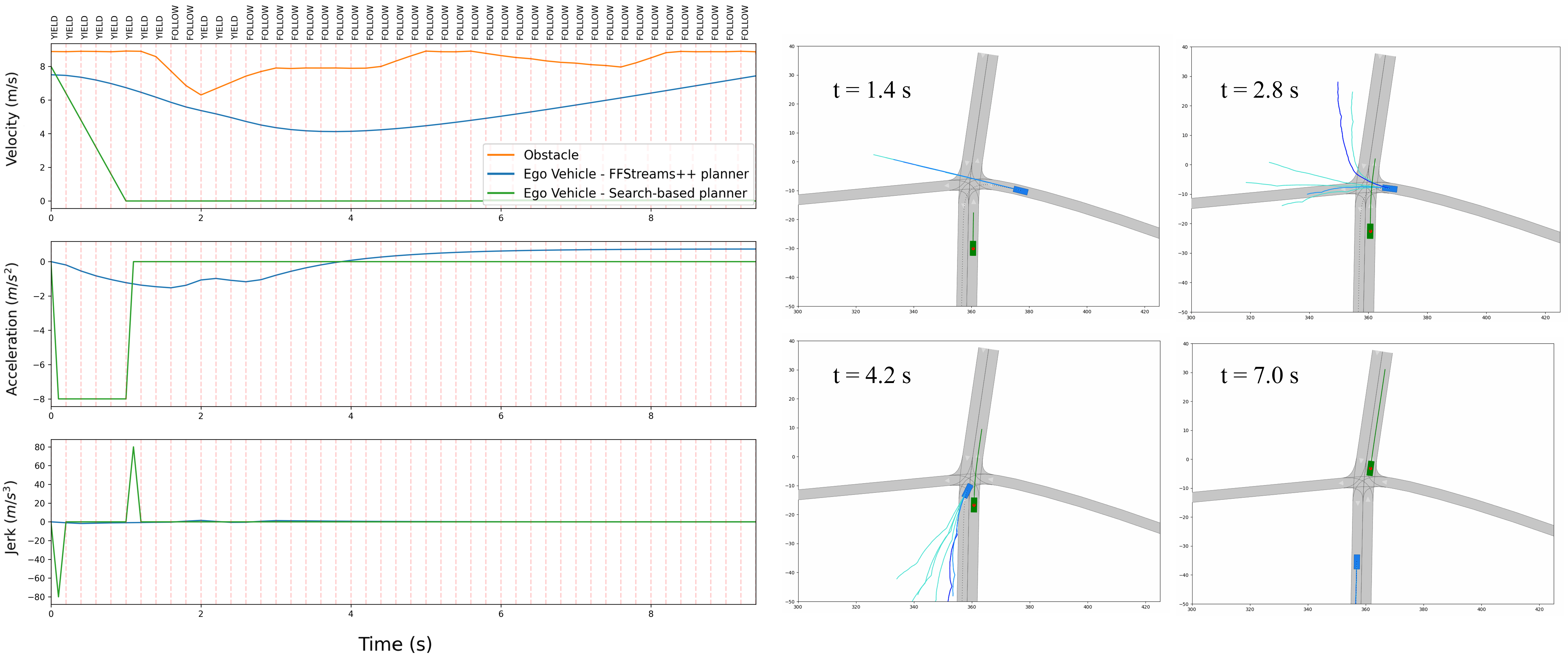}
 \caption{A successful experiment of passing the intersection planned by FFStreams++ planner and the velocity, acceleration and jerk profile of the planned trajectories as well as the decision at each timestep.}
 \label{fig_profiles_straight}
\end{figure*}

\section{Experiments and Results}
We tested the proposed framework for decision-making and motion planning of autonomous vehicles in complex intersection and overtaking scenarios to evaluate it. Multiple decision points exist at intersections, and there is a higher risk of potential conflicts. At crossing intersections with potential left turns, right turns, and straight paths, the autonomous vehicle should decide whether to proceed and pass the intersection or yield/stop for the other vehicles to pass. The decisions have to be accurate and safe, leading to a safe collision-free trajectory. In highway scenarios, where surrounding obstacles move at very high speeds and with the existence of slowly moving front obstacles, the planner should follow the normal driving behavior of overtaking when it is safe to overtake the front obstacle.

\begin{figure*}[t!]
 \centering
 \includegraphics[width=\textwidth]{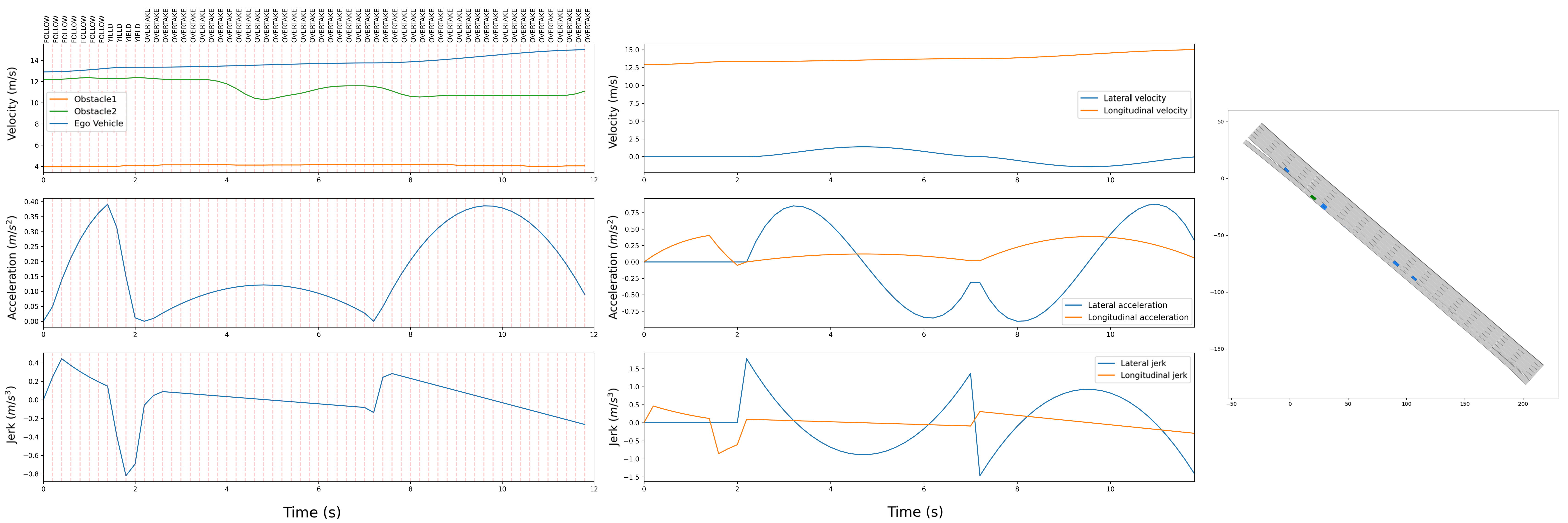}
 \caption{A successful experiment for overtaking maneuver. In the scenario, there is an obstacle, Obstacle1, moving at a lower speed ahead of the ego vehicle, and another three obstacles on the left neighbor lane. The neighbor obstacle behind the ego vehicle is Obstacle2 is moving at a high speed. The FFStreams++ planner makes decisions and plans a safe trajectory to overtake the front obstacle.}
 \label{fig_over_exp}
\end{figure*}

To evaluate the performance of the autonomous car, we use the metrics of safety and comfort for passengers, where the safety metric is related to the number of collision-free experiments, and comfort is defined by Occupant's Preference Metric (OPM) \cite{opm}, considering comfort based on the maximum lateral and longitudinal acceleration and jerk. We compared our method with the Search-Based planner in CommonRoad, which utilizes 2697 motion primitives and is based on the Search-Based Planning Library (SBPL) \cite{Likhachev-2018-109762}.

The experiments were carried out locally on an Ubuntu system with an Intel Core i7-10700F CPU: 16x2.90GHz computer with 32 GB RAM and an NVIDIA GeForce RTX 3060 Ti video card.  

\begin{table}
\caption{Results of Intersection simulations\label{intersection_results}}
\centering
\begin{tabular}{llll}
\toprule
{Method} & {Go straight}& {Left turn}&{OPM}  \\
{} & {Success rate}& {Success rate}&{}  \\
\midrule
{FFStreams++ planner}  & {$89\%$}  &{$84\%$}&{Normal Driver}\\
\hline
{Search-based planner}  &{$75\%$}  &{$53\%$} &{Aggressive Driver}  \\
\bottomrule
\end{tabular}
\end{table}

\begin{table}
\caption{Results of Highway simulations\label{intersection_results}}
\centering
\begin{tabular}{llll}
\toprule
{Method} & {Overtaking}& {Overtaking(after Waiting} &{OPM}  \\
{} & {Success rate}& {for Obstacle2 to pass)} &{}  \\
\midrule
{FFStreams++}  & {$92\%$}  &{$47\%$}&{Normal/Aggressive} \\
{planner}  & {}  &{}&{Driver} \\
\hline
{Search-based}  &{$34\%$}  &{$16\%$}&{Aggressive Driver}   \\
{planner}  & {}  &{}&{} \\
\bottomrule
\end{tabular}
\end{table}

\subsection{CommonRoad Benchmark}

CommonRoad is a valuable platform for evaluating and benchmarking the performance of autonomous driving algorithms. Its comprehensive and standardized set of scenarios covers a wide variety of scenarios, including urban intersections, highways, and rural roads, designed to test different aspects of decision-making and motion planning. 

We use two of the widely used CommonRoad's maps: an unsignalized intersection (DEU\_Nuremberg-39, Figure \ref{fig_intersection}) and a highway (USA\_US101-22, Figure \ref{fig_overtaking}). 
In addition, we run experiments in the CommonRoad simulator by randomly generating obstacles to cover
a wide range of challenging situations, including unprotected left turns at intersections, lane-keeping, and overtaking on a highway.

\subsection{Intersection scenarios}

We run the experiments in the CommonRoad simulator on the map DEU\_Nuremberg-39 (Figure \ref{fig_intersection}) with an unsignalized three-three way intersection. Starting from the initial position of the ego vehicle, shown in the picture, the ego (Green Car) vehicle has two possible maneuvers to perform: a Left Turn, where it performs a left turn from the south to the west and a Straight Path where it passes the intersection straight through the intersection. On the other hand, the obstacle (Blue Car) performs a left turn from east to south at different possible velocities and starting from a random initial position. 

In the intersection scenarios, the obstacle has an initial random position on the lane and moves at a high speed following a speed and acceleration profile taken from CommonRoad's realistic scenario "DEU\_Nuremberg-39\_5\_T-1". The ego vehicle has an initial position of [360.51,-30.79] and moves at a random initial speed [3.5 m/s, 11.5 m/s]. After conducting 100 experiments for two types of scenarios(left turn and go straight), in 89.00\% of the experiments of the "Go straight" scenario, the ego vehicle successfully drove straight and passed the intersection. In 84.00\% of the experiments of the "left turn" scenario, the ego vehicle successfully performed an unprotected left turn. On the other hand, the search-based planner had a success rate of 75.00\% in the experiments of the "Go straight" scenario and only 53.00\% in the experiments of the "Left turn" scenario. The results demonstrate the superior performance of FFStreams++ over search-based planners. After evaluating the performance of the two methods on the OPM metric, FFStreams++ trajectories were classified as a Normal Driver, while Search-based trajectories were classified as Aggressive Driver, proving the better performance of FFStreams and its closeness to human-like driving behaviour.

Figure~\ref{fig_profiles_left} shows a successful experiment with an unprotected left turn. In this scenario, FFStreams++ yields while approaching the intersection, and with the clearer intent of the surrounding vehicle, represented by its predicted trajectory, the FFStreams++ planner takes the decision to accelerate and perform the left turn. Figure also shows the velocity, acceleration, and jerk profiles as well as decisions planned by FFStreams++ planner, which adhere to acceleration constraints and have optimal jerks, leading to a comfortable ride for passengers. Search-based planners, on the other hand, failed at planning a future safe trajectory.

Figure~\ref{fig_profiles_straight} shows a successful experiment of passing intersections. In this scenario, FFStreams++ yields while approaching the intersection, and with the clearer intent of the surrounding vehicle, represented by its predicted trajectory, the FFStreams++ planner makes the decision to accelerate and pass the intersection. Figure also shows the velocity, acceleration, and jerk profiles as well as decisions planned by FFStreams++ planner, which adhere to acceleration constraints and have optimal jerks, leading to a comfortable ride for passengers. The figure also shows the planned trajectory by the Search-based planner, which includes stopping the ego vehicle from avoiding a collision. This behavior is far from normal driving behavior. On the other hand, the FFStreams++ trajectory is smooth, and decisions are close to the real driving behavior of humans.

\subsection{Highway scenarios}

We run the experiments in the CommonRoad simulator on the map USA\_US101-22 (Figure \ref{fig_overtaking}). The scenario consists of four obstacles: 1). a front obstacle on the same lane as the ego vehicle, moving at a low speed of $3.96 m/s$ and following a velocity and acceleration profiles extracted from one of CommonRoad's scenarios. 2). an obstacle on the left neighbor lane, existing behind the ego vehicle at a random distance of $[25m, 50m]$ and moving at a random speed of $[10m/s, 14m/s]$, and following an acceleration profile, extracted from one of CommonRoad's scenarios. 3) an obstacle on the left neighbor lane, way ahead of the ego vehicle, out of the region of interest of the ego vehicle. 4) an obstacle on the left neighbor lane, way ahead of the ego vehicle, out of the region of interest of the ego vehicle. 
In the scenario, the ego vehicle's initial position on XY is $[22.041,-18.688]$, its initial random velocity is $[12m/s, 14m/s]$, and it has zero initial acceleration.

After conducting 100 experiments with overtaking scenarios, in 92\% of the experiments, the ego vehicle successfully completed the front obstacle overtaking planned by the FFStreams++ planner. In 47\% out of 92\% successful overtaking experiments, the FFStreams++ planner has planned an overtaking maneuver after waiting for the neighbor obstacle Obstacle2 to pass, which reflects the safety criteria of the planner. On the other hand, the search-based planner has successfully performed overtaking only in 34\% of the experiments, and in 16\% of the search-based planner has planned to overtake after waiting for the neighbor obstacle to pass. After evaluating the performance of the two methods on OPM metric in Highway scenarios, part of FFStreams++ trajectories was classified as a Normal Driver, and part of the trajectories was classified as Aggressive Driver, while Search-based trajectories were all classified as Aggressive Driver, proving the better performance of FFStreams and a closer behavior to a human driver's behavior.

Figure \ref{fig_over_exp} demonstrates a highly critical overtaking experiment in a scenario with obstacles moving at a high speed. FFStreams++ has successfully planned decisions and a safe trajectory for performing overtaking maneuvers. The profiles of the planned absolute velocity, acceleration, and jerk are demonstrated, as well as lateral and longitudinal velocity, acceleration, and jerk. The planned trajectory by FFStreams++ has low acceleration values respecting the acceleration constraint and low optimized jerk values, leading to a smooth, comfortable trajectory for passengers. Search-based planners, on the other hand, failed at planning a future safe trajectory.

\section{Analysis of Experimental Results}

In intersection scenarios, our experimental results indicate that the failure rate is notably higher in left-turn scenarios compared to go-straight scenarios, and both are higher than the failure rate in highway scenarios. This can be attributed to the inherent challenges in accurately predicting the trajectories of oncoming vehicles at intersections. The current prediction model relies heavily on the historical states of obstacles and the polygons of the road. However, it does not explicitly incorporate intersection driving rules, such as yielding before the intersection to observe surrounding vehicle states, which are important for accurate trajectory prediction in such complex scenarios. Additionally, the model tends to prioritize historical data over road geometry, which, as illustrated in Figures \ref{fig_profiles_left} and \ref{fig_profiles_straight}, sometimes leads to predicted trajectories that deviate from the road boundaries.

This limitation significantly impacts the decision-making process, particularly when there is a delay in anticipating that an oncoming vehicle will decelerate before the intersection. Such delays can result in the planning framework making a series of hesitant decisions, which can compromise the smoothness of the overall decision-making process. For instance, the framework might alternate between decisions to yield(decelerate) and to follow(accelerate) the target speed over multiple planning cycles before finally committing to a safe maneuver. As demonstrated in figure \ref{fig_profiles_straight} making the decision of Follow for times $[1.6s, 1.8s]$, and Yield for times $[2.0s, 2.2s, 2.4s]$, then Follow decision consistently till the end of the scenario. Despite these challenges, the framework's ability to replan within less than 200 milliseconds allows it to adapt to changes in predictions, ultimately passing intersections safely following planned trajectories.

The proposed FFStreams++ framework demonstrates a strong capacity to adapt and generate smooth trajectories, even when faced with frequent decision reversals between yielding and following (accelerating). This adaptability is achieved through optimized trajectories that minimize jerk and adhere to strict constraints on maximum acceleration, ensuring a smooth and safe response in dynamically changing environments.

In highway scenarios, the predicted trajectories of obstacles are significantly more accurate. This accuracy is largely due to the fact that obstacles on highways tend to maintain their heading unless they explicitly intend to change lanes. The lower probability of sudden heading changes simplifies the prediction task, leading to more reliable planning outcomes.

Regarding failure rates, the proposed framework exhibits a failure rate of 8\% in overtaking scenarios and 11\% and 16\% in intersection scenarios. Overall, the failure rate of the framework ranges from 9\% to 16\%, which we consider acceptable given the challenging and complex nature of the scenarios, including highway overtaking and navigating unsignalized intersections with high-risk factors.

The robustness of the framework across different vehicle kinematics and types in various scenarios is guaranteed through several key features:
\begin{itemize}
\item Collision-checking: This is performed by simulating each future driving timestep, where the 2D projections of vehicles on predicted trajectories are checked against the 2D projection of the ego vehicle on the sampled trajectories using the CommonRoad Drivability Checker library.
\item Parameter Flexibility: The Frenet and Jerk optimization parameters can be adjusted to accommodate the maximum acceleration and curvature values specific to the target ego vehicle, ensuring the framework’s adaptability.
\item Consistent Performance: The framework consistently delivers reliable performance across diverse scenarios, including both highway and urban environments.
\end{itemize}
Overall, the proposed framework demonstrates a high degree of reliability and adaptability, making it well-suited for the demands of autonomous driving in both urban intersections and highway scenarios.

\section{Conclusion and Future Work}
We introduced the FFStreams++ framework, an advanced approach for integrated decision-making and motion planning in autonomous driving systems. FFStreams++ combines sampling and search-based methods by iteratively sampling maneuver-specific trajectories to update the initial PDDL problem's state using Streams and searching for an optimal plan using FastForward heuristic search. It takes advantage of integrated decision-making and motion planning to find an optimal plan with only feasible trajectories.

By integrating a Query-Centric network (QCNet) for trajectory prediction, FFStreams++ accurately predicts the movements of surrounding obstacles, including their accelerations, enhancing the system's decision-making capabilities. Our experimental evaluations on the CommonRoad simulation framework demonstrate the effectiveness of the FFStreams++ planner in executing complex maneuvers such as unprotected left turns and overtaking. The results demonstrate that FFStreams++ performs consistently, safely, and adapts well to dynamic and unpredictable environments in both highway and urban scenarios. This work contributes to the advancement of autonomous vehicle technology by providing a robust and reliable framework for motion planning and trajectory prediction.

In future research, we intend to enhance the precision of our prediction model and extend the capabilities of our framework to handle a broader range of driving maneuvers, including highway merging, exiting, and intersection priority management. This will involve integrating traffic light and crosswalk semantics into our framework to enable more nuanced and context-aware decision-making in complex traffic scenarios. By broadening the scope of FFStreams++ and improving prediction accuracy, we aim to enhance the versatility and robustness of autonomous driving systems, ensuring their safe operation across an even wider array of driving conditions.


\bibliographystyle{unsrtnat}
\bibliography{mybibfile}  






\end{document}